\documentclass{article}
\usepackage{graphicx} % Required for inserting images
\usepackage{amssymb}
\usepackage{amsmath}
\usepackage{algorithm}
\usepackage{algorithmic}
\usepackage{graphicx}
\usepackage{subfigure}
\usepackage{multirow}
\usepackage{tabularx}
\usepackage{breqn}

\title{SIL-RRT*: Learning Sampling Distribution through Self Imitation Learning}
\author{%
Xuzhe Dang \\
\textit{dangxuzh@fel.cvut.cz} \\
\textit{Czech Technical University in Prague} \\
\and
Stefan Edelkamp \\
\textit{stefan.edelkamp@aic.fel.cvut.cz}\\
\textit{Czech Technical University in Prague}
}

\begin{document}

\maketitle

\begin{abstract}
Efficiently finding safe and feasible trajectories for mobile objects is a critical field in robotics and computer science. In this paper, we propose SIL-RRT*, a novel learning-based motion planning algorithm that extends the RRT* algorithm by using a deep neural network to predict a distribution for sampling at each iteration. We evaluate SIL-RRT* on various 2D and 3D environments and establish that it can efficiently solve high-dimensional motion planning problems with fewer samples than traditional sampling-based algorithms. Moreover, SIL-RRT* is able to scale to more complex environments, making it a promising approach for solving challenging robotic motion planning problems. 
\end{abstract}

\section{Introduction}

Motion planning is a crucial field of study in robotics and computer science that focuses on finding a feasible and safe trajectory for a robot to achieve a desired goal. It involves determining a sequence of actions that will guide the robot from its initial state to the target, while avoiding collisions and satisfying various constraints, such as kinematic limitations, time constraints, and performance criteria. The significance of motion planning lies in its ability to enable robots to interact safely with their environment and carry out various tasks autonomously.

As robots become more widely used and their applications increasingly complex, the motion planning problem demands algorithms that are both computationally tractable and efficient. This has led to the development of sampling-based algorithms, such as Probabilistic Roadmaps (PRM)~\cite{kavraki1996probabilistic}~, Rapidly-exploring Random Trees (RRT)~\cite{lavalle2001randomized}~, and RRT*~\cite{karaman2011sampling}~. These algorithms typically employ a uniform sampler, but when the dimension of the environment increases, the number of samples needed to find a feasible solution may also raise. To address this sampling efficiency problem, some researchers have introduced heuristic biased samplers, such as BIT*~\cite{gammell2015batch}~ and Informed RRT*~\cite{gammell2014informed}~.

In this study, we introduce SIL-RRT*, a novel approach that enhances the Rapidly-exploring Random Tree Star (RRT*) algorithm through the utilization of deep learning techniques. Unlike traditional methods that rely on Convolutional Neural Networks (CNNs) or Graph Neural Networks (GNNs), we adopt a Transformer-based architecture to capture the relationships between the state space and paths. This choice of architecture allows for greater flexibility and extensibility, enabling SIL-RRT* to handle motion planning tasks across different dimensions.

\begin{figure*}[t]
\centering
	\includegraphics[width=0.3\linewidth]{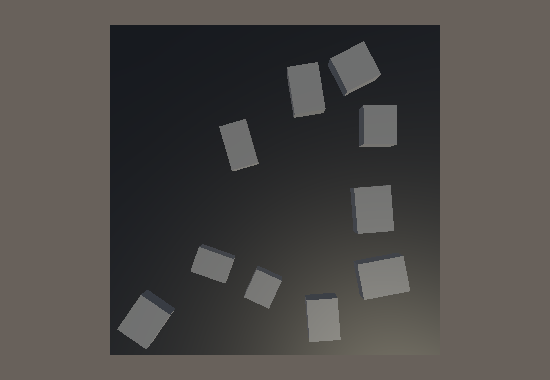}\hfill
	\includegraphics[width=0.3\linewidth]{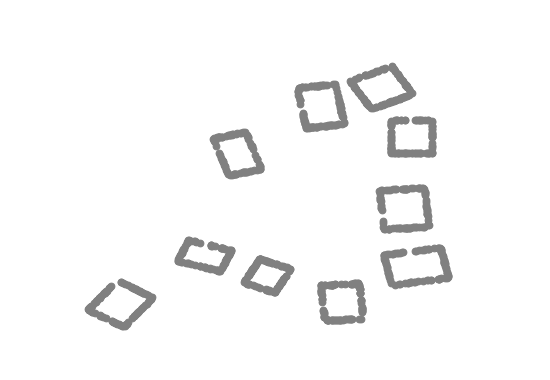}\hfill
	\includegraphics[width=0.3\linewidth]{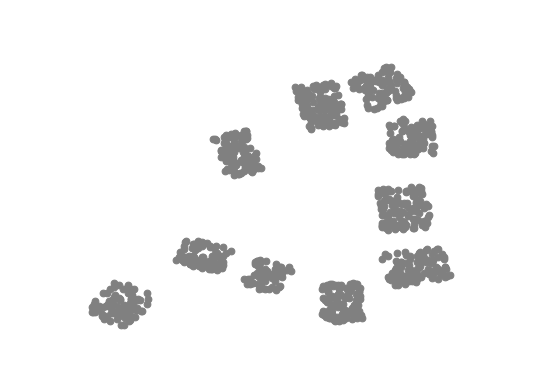}
\caption{Illustration of a 2D state space scenario and two contrasting methods for representing the state space using point clouds. In our experiment, we opted to sample point clouds from the surface of obstacles rather than their interior regions.}
\label{fig:example_scenario}
\end{figure*}

In SIL-RRT*, the state space is represented by a list of point clouds sampled on the surface of obstacles, providing a comprehensive depiction of the environment, as shown in as Figure~\ref{fig:example_scenario}. Indeed, the technique we employed in SIL-RRT* for representing obstacles using point clouds sampled from their surfaces is similar to the approach utilized in the work by Strudel et al. (2021)~\cite{strudel2021learning}. This technique has been shown to be effective in representing obstacles in various dimensions. We trained SIL-RRT* using a dataset collected within our Unity-based environment, enabling the algorithm to learn from a diverse range of scenarios.

To further enhance SIL-RRT*, we incorporated a fine-tuning technique called weighted self-imitation learning (WSIL), which dynamically selects high-quality solutions from a fixed-size buffer. This approach eliminates the requirement of collecting near-optimal demonstrations by reusing successful records during the training process. This technique is particularly valuable in high-dimensional environments, where obtaining such demonstrations can be difficult.

In our experiments, extensive evaluations in diverse motion planning scenarios have demonstrated the effectiveness of SIL-RRT*. The algorithm significantly reduces the number of samples required to solve new planning tasks, effectively leveraging prior experiences for improved performance in both 2D and 3D environments. Overall, our study contributes a cutting-edge motion planning approach, SIL-RRT*, that combines deep learning, self-imitation learning, and a Transformer-based architecture to enhance the RRT* algorithm for high-dimensional motion planning problems.

\section{Related Work}

The idea of improving the sample generation strategy of RRT has been widely researched in the field of motion planning~\cite{gammell2014informed,islam2012rrt}. Although these studies have focused on enhancing the quality of paths, they still have limitations in terms of computational efficiency.

Recently, there has been a significant increase in the popularity of learning-based motion planning algorithms, which hold the promise of solving problems more efficiently. These methods can be broadly classified into three categories based on how the Deep Neural Network (DNN) is trained: Supervised,  Imitation, or Reinforcement Learning.

Supervised Learning-Based: The MPNet algorithm~\cite{qureshi2019motion} employs a DNN trained through supervised learning to directly predict the next state of a feasible solution. The NR-RRT approach~\cite{meng2022nr} leverages a DNN, trained with image-based risk contours maps and states of the current and goal, as a sampler for the RRT algorithm. Other studies~\cite{ichter2018learning, khan2020graph, kumar2019lego, baldoni2022leveraging} utilize a conditional variational autoencoder to learn the distribution of feasible solutions. 
A convolutional block attention generative adversarial network for sampling-based path planning has been proposed by~\cite{sagar2023cbaganrrt}.
However, these supervised learning-based algorithms generally require a large amount of training data, and the resulting models may lack the ability to adapt to unseen environments.

Imitation Learning-Based: Imitation learning is a machine learning approach where an agent learns to execute a task by observing and imitating the actions of an expert or teacher. The work of~\cite{luo2021self} and 
\cite{zhang2022learning} utilized imitation learning in their algorithms to control a robot arm. Deep-RRT*~\cite{dang2022deep} extends the RRT* algorithm by incorporating a Deep Neural Network (DNN) to operate in 2D environments. Just like supervised learning, the performance of imitation learning is tightly tied to the quality of expert demonstrations and may struggle with adapting to new, unseen environments. To address these challenges, both studies employ the DAgger algorithm~\cite{arxiv.1801.06503}.

Reinforcement Learning-Based: Building upon the success of applying reinforcement learning in Atari games, numerous efforts have been made to train models for solving motion planning problems using reinforcement learning. The RL-RRT algorithm~\cite{chiang2019rl} utilizes a model trained through reinforcement learning to minimize the number of samples needed. On the other hand, the NEXT approach~\cite{chen2019learning} combines Upper Confidence Bound for Trees (UCT) with the RRT algorithm, utilizing both a policy and a value network.

\section{Preliminaries}
\subsection{Problem Statement}
The state space, $S$, is defined as a subset of $\mathbb{R}^n$. The obstacle state space, $S_{obs}$, and free state space, $S_{free}$, are defined as subsets of $S$. The initial state, $S_{init}$, is defined as an element of $S_{free}$, and the goal region, $S_{goal}$, is defined as a subset of $S_{free}$. A collision-free path, $\tau$, is defined as a continuous mapping from $[0,1]$ to $S_{free}$ such that $\tau(0)=S_{init}$ and $\tau(1)\in S_{goal}$. The set of all collision-free paths is defined as $T$. The cost function, $c(\cdot)$, is used to evaluate paths and the optimal motion planning problem is to find the path, $\tau^*$, with the minimum cost among all paths in $T$.

	\begin{equation}
		\tau^{*} = \operatorname{argmin}_{\tau \in T} c(\tau)
	\end{equation}

\subsection{Transformer}

Transformer~\cite{vaswani2017attention} is a powerful neural network architecture that is specially designed to process sequential data, such as text, speech, and time series. The architecture consists of stacks of self-attention layers with residual connections, allowing the model to selectively focus on relevant parts of the input sequence.
In a typical Transformer model, the sequence of input vectors $\{x_1, x_2, ..., x_n\}$ is transformed into key vectors $\{k_1, k_2, ..., k_n\}$, query vectors $\{q_1, q_2, ..., q_n\}$, and value vectors $\{v_1, v_2, ..., v_n\}$ through linear transformations. The weighted sum of value vectors is computed using the self-attention mechanism

\begin{equation}
a = \mbox{softmax}\left(\frac{QK^T}{\sqrt{d_k}}\right)V
\end{equation}

where $Q$ denotes the query vectors, $K$ the key vectors, and $V$ the value vectors, with $d_k$ being the dimensionality of the key vectors.

Attention mechanisms have proven to be effective in handling long-range dependencies within input data by dynamically selecting relevant information based on context. As a result, they have been utilized in fields beyond NLP, such as computer vision\cite{dosovitskiy2020image} and reinforcement learning~\cite{chen2021decision}.

\begin{figure*}[t]
  \centering
  \begin{minipage}[b]{0.45\textwidth}
    \centering
    \includegraphics[width=\textwidth]{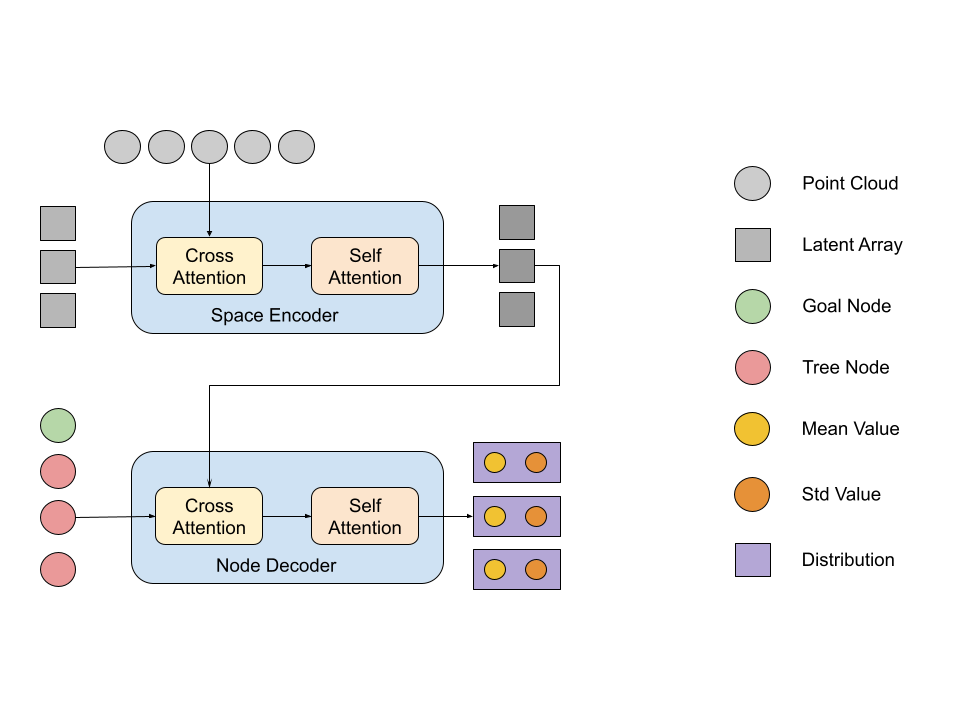}
    \caption{Architecture of Sampler Model}
    \label{fig:Sampler Model}
  \end{minipage}
  \hfill
  \begin{minipage}[b]{0.45\textwidth}
    \centering
    \includegraphics[width=\textwidth]{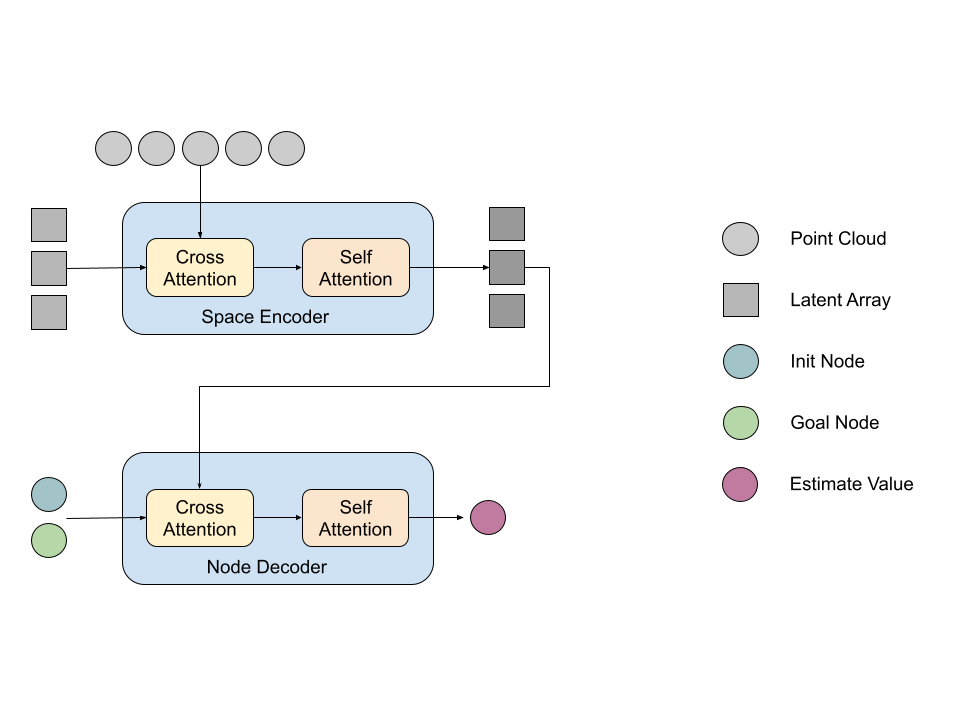}
    \caption{Architecture of Estimator}
    \label{fig:Estimator}
  \end{minipage}
\end{figure*}

In SIL-RRT*, we have implemented an attention mechanism to effectively process the relationships among obstacles, goals, and the expanded tree. This mechanism enables us to generate a distribution that guides the tree expansion process, improving the efficiency and effectiveness of the algorithm. To implement this attention mechanism, we have adapted the powerful Transformer architecture, which we will delve into further in subsequent discussions.

The Transformer architecture's ability to capture global dependencies and establish long-range relationships proves advantageous for managing the complexities inherent in motion planning tasks. SIL-RRT* utilizes this capability to facilitate efficient and effective tree expansion, resulting in high-quality solutions in challenging environments.

\subsection{Imitation Learning}
Imitation learning is a machine learning technique where an agent learns to perform a task by imitating the actions of an expert. The goal of imitation learning is to enable the agent to learn from a set of expert demonstrations, rather than learning through trial and error in the environment. Imitation learning is particularly useful in cases where defining a reward function for the task is difficult or where direct interaction with the environment during learning is challenging or risky.

Imitation learning has been applied to various tasks, such as robotics~\cite{fang2019survey}, autonomous driving~\cite{zhang2016query,bhattacharyya2022modeling}, and game playing~\cite{reddy2019sqil, oh2018self}. Many studies have explored the use of imitation learning to train agents in complex tasks, often with promising results.

We enhance our model using self-imitation learning, drawing from solutions created by RRT* and SIL-RRT*. These act as 'experts', contributing to the training data, fostering diverse learning experiences. This bolsters the model's robustness and reliability by imitating its effective solutions.

\section{Method}
\begin{figure*}[t]
  \centering
  \begin{tabular}{cccc}
    \includegraphics[width=0.22\textwidth]{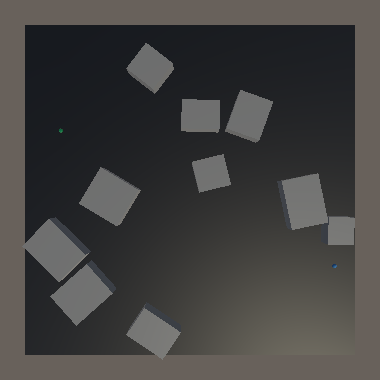} &
    \includegraphics[width=0.22\textwidth]{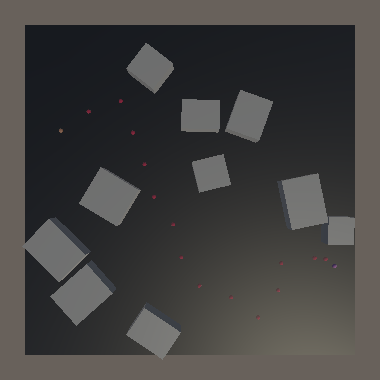} &
    \includegraphics[width=0.22\textwidth]{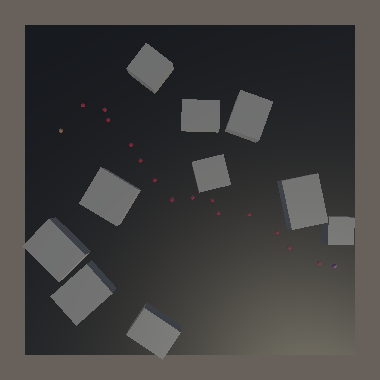} &
    \includegraphics[width=0.22\textwidth]{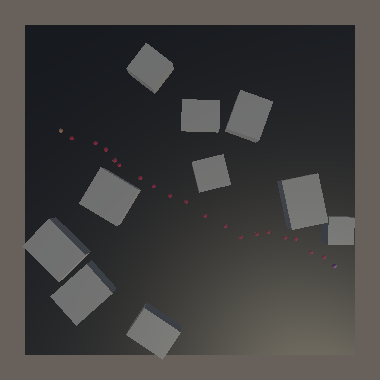} \\
    \includegraphics[width=0.22\textwidth]{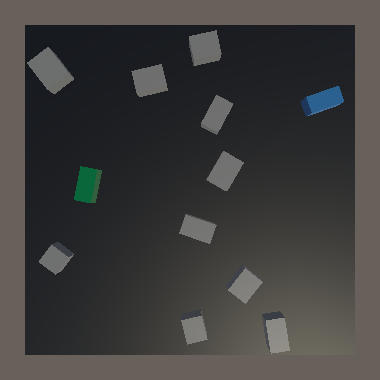} &
    \includegraphics[width=0.22\textwidth]{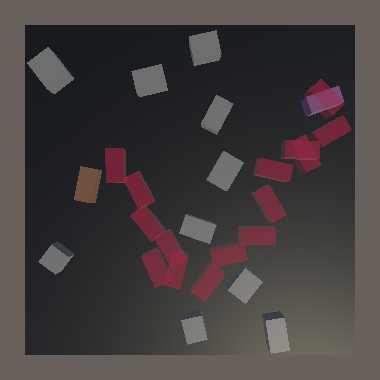} &
    \includegraphics[width=0.22\textwidth]{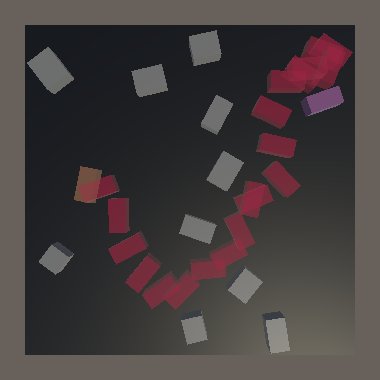} &
    \includegraphics[width=0.22\textwidth]{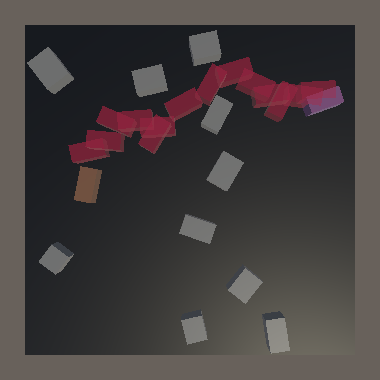} \\
    \includegraphics[width=0.22\textwidth]{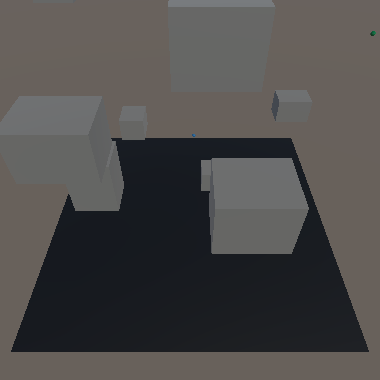} &
    \includegraphics[width=0.22\textwidth]{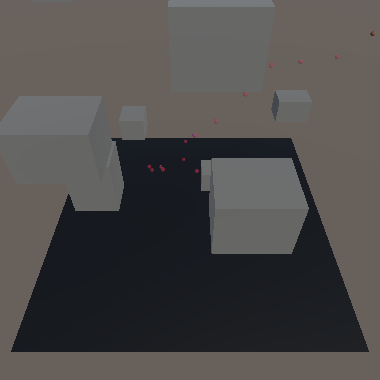} &
    \includegraphics[width=0.22\textwidth]{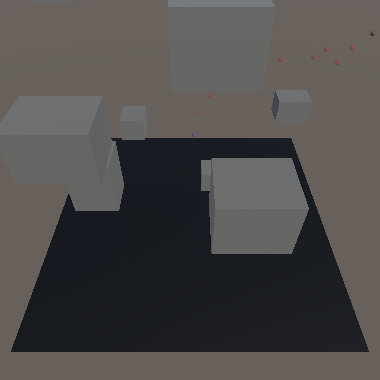} &
    \includegraphics[width=0.22\textwidth]{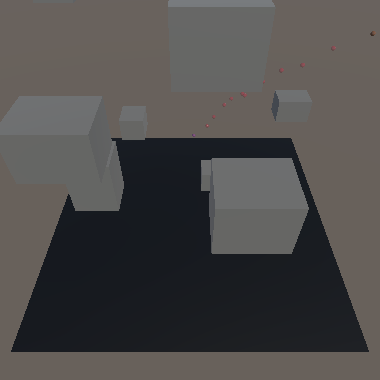} \\
    \includegraphics[width=0.22\textwidth]{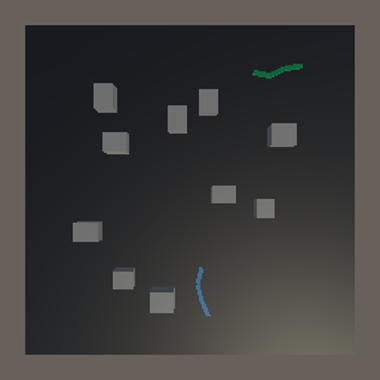} &
    \includegraphics[width=0.22\textwidth]{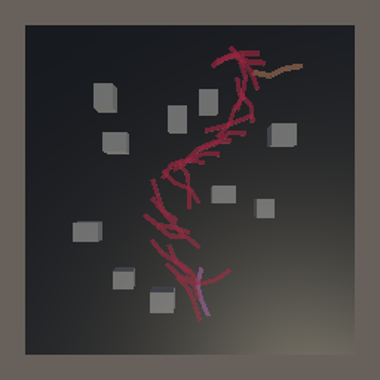} &
    \includegraphics[width=0.22\textwidth]{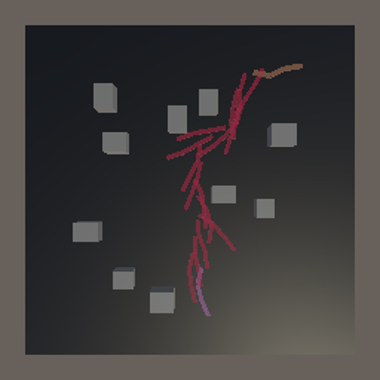} &
    \includegraphics[width=0.22\textwidth]{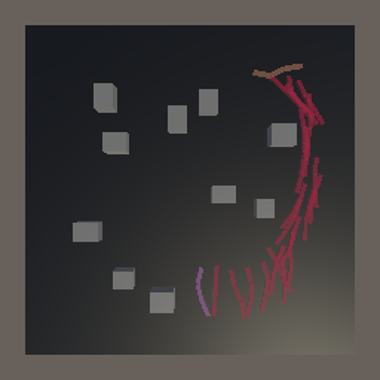} \\
  \end{tabular}
  \caption{Examples of paths found by RRT*, SIL-RRT* w/o and with WSIL algorithms in various scenarios. Left column shows the scenario and 
  the following ones the trajectories found, which are best
  visualized magnified and in colors on screen.}
  \label{fig:demos}
\end{figure*}

Imitation learning is a machine learning technique where an agent learns to perform a task by imitating the actions of an expert. The goal of imitation learning is to enable the agent to learn from a set of expert demonstrations, rather than learning through trial and error in the environment. Imitation learning is particularly useful in cases where defining a reward function for the task is difficult or where direct interaction with the environment during learning is challenging or risky.

Imitation learning has been applied to various tasks, such as robotics~\cite{fang2019survey}, autonomous driving~\cite{zhang2016query,bhattacharyya2022modeling}, and game playing~\cite{reddy2019sqil, oh2018self}. Many studies have explored the use of imitation learning to train agents in complex tasks, often with promising results.

We enhance our model using self-imitation learning, drawing from solutions created by RRT* and SIL-RRT*. These act as 'experts', contributing to the training data, fostering diverse learning experiences. This bolsters the model's robustness and reliability by imitating its effective solutions.

\subsection{Sampler Model Architecture}

In the SIL-RRT* framework, we employ a Transformer architecture to efficiently process point cloud data representing obstacles, subsequently generating a target sequence for navigation, as depicted in Figure~\ref{fig:Sampler Model}. This architecture comprises two primary components: the State Space Encoder and the Decoder. The State Space Encoder transforms the raw point cloud data into a fixed-length vector representation, effectively capturing the spatial distribution of obstacles within the environment. The Decoder then uses this representation to methodically encode the tree's growth, enhancing the state space exploration efficiency while considering spatial constraints imposed by obstacles.

State Space Encoder: Central to the SIL-RRT* framework, this encoder processes the point cloud data sequence into a standardized, fixed-length representation. Given the computational intensity associated with traditional transformer architectures, we employ the Perceiver model \cite{jaegle2021perceiver}, which adeptly handles the complexity and scale of input data. When presented with point cloud obstacles, noted as $p = {p_1, p_2, \ldots, p_n}$, with $n$ indicating the sequence's length, the encoder utilizes a cross-attention mechanism to project $p$ onto a compact, fixed-length latent representation $Z_{p}$. This is subsequently refined via a self-attention layer, enhancing its information content and structural integrity for downstream processing.

Node Decoder: Utilizing the Perceiver-IO architecture \cite{jaegle2021perceiverio}, the Decoder processes the input sequence of tree nodes, represented as $x = {goal, x_1, x_2, \dots, x_t}$, where $t$ signifies the sequence's total node count. A causal attention mask ensures no future node information is encoded, maintaining the sequence's temporal integrity. Mirroring techniques from Decision Transformers \cite{chen2021decision}, the model focuses only on the last five nodes of the sequence, balancing computational efficiency with predictive accuracy.

The decoding phase begins with a cross-attention layer that maps the latent array $Z_p$, derived from the State Space Encoder, onto the latent features $Z_x$ corresponding to the sequence $x$. This ensures the decoded features faithfully represent the encoded spatial information. A subsequent self-attention layer further refines these features, enhancing their representational utility. The process concludes with two Multilayer Perceptron (MLP) layers that calculate mean(
$\mu$) and standard deviation ($\sigma$). These parameters define a multivariate normal distribution, from which samples are drawn to construct the RRT*, enabling the generation of path samples that are both diverse and contextually informed by the environment's spatial layout.

In this study, we utilize feasible paths, denoted as $\tau = {x_0, x_1, x_2, \dots, x_t}$, sourced from planning algorithms or expert demonstrations, to guide our model. To enhance the diversity of our dataset, we introduce a novel data augmentation technique that involves random reversals of these paths. This approach substantially increases the variety of training data available. We aim to optimize our model by minimizing the negative log-likelihood loss, employing a multivariate normal distribution to accurately model the underlying data distribution. This precise estimation of data characteristics is critical for effective model training.

\begin{dmath}
L_{sampler} = - \frac{1}{B} \sum^{N}_{i=1}\log\pi_{\theta}(x_{i}\mid p, g, x_{i-1})
\end{dmath}

The SIL-RRT* algorithm operates through a structured multi-stage process, starting with the encoding of point clouds and goal positions, followed by decoding to form a tree structure, and concluding with the generation of samples essential for constructing the Rapidly-exploring Random Tree (RRT). This method effectively manages the intricate interactions between point cloud data and tree nodes, thus enabling the algorithm to deliver comprehensive navigation solutions for complex environments.

During experimental trials, it was observed that the sampling mechanism occasionally struggles near goal states. This results in the generation of excessive samples in these areas and, at times, a failure to identify feasible paths. To address these challenges, we integrated the BiRRT* algorithm, as detailed by Jordan and Pere \cite{jordan2013optimal}, which utilizes alternating forward and backward tree expansions. This strategy has proven effective in overcoming the observed issues, significantly improving the algorithm's ability to find feasible paths through complex spaces.

\subsection{Weighted Self Imitation Learning}
In both supervised and imitation learning frameworks, the quality of expert demonstrations is paramount; however, acquiring optimal or near-optimal demonstrations presents considerable challenges. Within our proposed SIL-RRT* framework, these challenges are exacerbated due to the inherent properties of Rapidly-exploring Random Trees (RRT), which prioritize solution feasibility over optimality. Consequently, our dataset inevitably comprises a mix of optimal and sub-optimal trajectories. This heterogeneity complicates the task of identifying and filtering out low-quality trajectories, potentially diminishing the performance and efficacy of the trained sampler model.

Addressing the need for high-quality demonstrations in imitation learning, several innovative methodologies have been developed. Self-Imitation Learning (SIL) as proposed by Oh et al.\cite{oh2018self}, focuses on replicating an agent's past successful actions, effectively ignoring less successful ones. Complementary strategies such as 2IWIL \cite{wu2019imitation} and SAIL \cite{wang2021robust} incorporate a quality weighting mechanism into Generative Adversarial Imitation Learning (GAIL) \cite{ho2016generative}, enhancing the learning process from a diverse array of expert demonstrations. Additionally, T-REX \cite{brown2019extrapolating} develops a reward function that evaluates demonstrations based on their quality and informativeness, prioritizing the most valuable examples.

To discern high-quality demonstrations within the SIL-RRT* framework, we adopt a similar approach by using the length of each trajectory as a measure of its quality. We deploy a Deep Neural Network (DNN) as a predictive estimator to forecast the path length produced by the SIL-RRT* using the current sampler model, as illustrated in Fig.~\ref{fig:Estimator}. We introduce a threshold, denoted as $K$, to distinguish between superior and inferior demonstrations, based on the discrepancy between the actual and predicted path lengths. Each solution in our dataset is then weighted according to this measure, as specified by Equation (4):

\begin{dmath}
w = \frac{1}{\left(1+e^{(C_{real}-C_{est}-K)}\right)}
\end{dmath}

This methodological innovation aims to enhance the fidelity and utility of our dataset, thereby improving the overall performance of the training model.

\begin{table*}[t]
  \centering
  \caption{Comparison of SIL-RRT* with RRT* and MPNetNR with the simple scenarios.}
  \label{tab:comparison_simple}
  \resizebox{\textwidth}{!}{%
  \begin{tabular}{|l|c|c|c|c|c|}
    \hline
    \multirow{2}{*}{\textbf{Environment}} & \textbf{Metrics} & \textbf{SIL-RRT* w/o WSIL} & \textbf{SIL-RRT* with  WSIL} & \textbf{RRT*} & \textbf{MPNetNR} \\ 
    \cline{2-6}
    & \textbf{Success Rate} & $99.30\%$ & $99.13\%$ & $71.60\%$ & $44.93\%$ \\ 
    \cline{2-6}
    2D & \textbf{Average Samples Generated} & $22.53\pm18.05$ & $23.93\pm19.09$ & $98.26\pm60.47$ & $-$ \\ 
    \cline{2-6}
    & \textbf{Average Path Length} & $17.42\pm6.56$ & $17.01\pm6.52$ & $19.64\pm7.01$ & $28.68\pm15.56$ \\ 
    \cline{2-6}
    & \textbf{Average Time} & $0.24\pm0.14$ s & $0.26\pm0.15$ s & $0.30\pm0.16$ s & $0.15\pm0.05$ s \\
    \hline
    \multirow{2}{*}{Rigid Body} & \textbf{Success Rate} & $98.23\%$ & $98.06\%$ & $74.70\%$ & $49.53\%$ \\ 
    \cline{2-6}
    & \textbf{Average Samples Generated} & $25.30\pm25.37$ & $26.65\pm27.55$ & $83.15\pm57.73$ & $-$ \\ 
    \cline{2-6}
    & \textbf{Average Path Length} & $15.94\pm6.45$ & $15.58\pm6.33$ & $17.59\pm6.43$ & $27.15\pm15.44$ \\ 
    \cline{2-6}
    & \textbf{Average Time} & $0.27\pm0.19$ s & $0.28\pm0.21$ s & $0.29\pm0.16$ s & $0.19\pm0.10$ s \\ 
    \hline
    \multirow{2}{*}{3D} & \textbf{Success Rate} & $100\%$ & $100\%$ & $17.20\%$ & $82.47\%$ \\ 
    \cline{2-6}
    & \textbf{Average Samples Generated} & $17.78\pm6.28$ & $17.34\pm6.29$ & $362.19\pm83.89$ & $-$ \\ 
    \cline{2-6}
    & \textbf{Average Path Length} & $20.43\pm4.95$ & $17.69\pm4.25$ & $27.25\pm6.42$ & $68.94\pm33.01$ \\ 
    \cline{2-6}
    & \textbf{Average Time} & $0.17\pm0.03$ s & $0.17\pm0.4$ s & $1.41\pm0.39$ s & $0.25\pm0.04$ s \\ 
    \hline
  \end{tabular}
  }
\end{table*}

During the initial phase of training, because estimator is not well-trained; thus, we set $K$ to a large value to include more training samples. As the training progresses, we gradually decrease the value of $K$ to prioritize higher-quality demonstrations. We update $K$ in fixed steps by dividing it by $\mu$.
This weight takes into account the difference between the length of the real path and the estimated path, with solutions that are shorter than predicted having a larger impact on our sampler model. During training,
we update the parameters of our length estimator using the MSE loss function:
\begin{dmath}
L_{estimator} = \frac{1}{2} \vert \vert\ C_{real}-C_{est}  \vert \vert\ ^{2}
\end{dmath}

To fine-tuning our sampler model, we updated the negative log-likelihood loss loss function that takes into account both the weight of each solution. The loss function is defined as follows:

\begin{dmath}
L_{sampler} = - \frac{1}{B} \sum^{N}_{i=1}\log\pi_{\theta}(x_{i}\mid p, g, x_{i-1}) \cdot w - H_{\theta}^{T}[x_{i}\mid p, g, x_{i_i}] \cdot \lambda
\end{dmath}

Here, $B$ represents the batch size, $N$ denotes the number of nodes within a solution, $g$ is the goal state, and $p$ and $x$ correspond to the point cloud vectors and tree nodes, respectively. To foster the exploration of novel paths by our sampler model, we focus on minimizing the Shannon entropy of the model's predicted distribution, symbolized as $H_{\theta}^{T}$. This loss function is pivotal for learning from high-quality demonstrations, as it aids in preventing the model from overfitting to solutions of inferior quality. The description of the algorithm can be found at reference \ref{algo:wsil}.

\begin{algorithm}
	\caption{Weighted Self Imitation Learning}
        \label{algo:wsil}
	\begin{algorithmic}
	\STATE Initialize $S = \{Scenario_1, Scenario_2, ..., Scenario_n\}$
	\STATE Initialize $D = []$
	\STATE $\epsilon \leftarrow 1$
	\FOR {$i\leftarrow 1$ to $n$}
	\STATE ${x_{init}, x_{goal}, x_{obs}} \leftarrow Sample(S)$
	\IF{$rand() > \epsilon$}
	\STATE $\tau \leftarrow \mbox{SIL-RRT*}(x_{init}, x_{goal}, x_{obs})$
	\ELSE
	\STATE $\tau \leftarrow \mbox{RRT}(x_{init}, x_{goal}, x_{obs})$
	\ENDIF
	\IF{$\tau \neq \varnothing$}
	\STATE $D \leftarrow \mbox{\em Add}(D, \tau)$
	\ENDIF
	\STATE Sample $({\tau, x_{init}, x_{goal}, x_{obs}})$ from $D$
	\STATE $C_{est} \leftarrow \mbox{\em Estimator}(x_{obs}, x_{init}, x_{goal})$
	\STATE $w \leftarrow \frac{1}{\left(1+e^{(C_{real}-C_{est}-K)}\right)}$
	\STATE $\theta_{policy} \leftarrow \theta_{policy} - \eta \nabla_{\theta_{policy}}L_{policy} $
	\STATE $\theta_{estimator} \leftarrow \theta_{estimator} - \eta \nabla_{\theta_{estimator}}L_{estimator} $
	\STATE $\epsilon \leftarrow \mbox{\em Anneal}(i)$
	\ENDFOR
	\end{algorithmic}
\end{algorithm}

\section{Experiment Results}

\subsection{Data Collection}

We\footnote{Refer to the supplementary material accompanying this paper for videos and more in-depth studies.} utilized Unity to enhance the visualization and in- teraction of SIL-RRT*. To enable a smooth data exchange between Unity and external processes, we implemented a socket communicator on both the Unity and Python sides, allowing an effective training of SIL-RRT* through the exposure of registered functions. To ensure the versatility and robustness of SIL-RRT*, we have designed and generated three distinct environments: 2D, rigid body, and 3D. Each environment consists of 100 unique workspaces, and within each workspace, we have created 50 different scenarios.

\begin{table*}[t]
  \centering
  \caption{Comparison of SIL-RRT* with the complex scenarios.}
  \label{tab:comparison_complex}
  \resizebox{\textwidth}{!}{%
  \begin{tabular}{|l|c|c|c|}
    \hline
    \multirow{2}{*}{\textbf{Environment}} & \textbf{Metrics} & \textbf{SIL-RRT* w/o WSIL} & \textbf{SIL-RRT* with WSIL} \\ 
    \cline{2-4}
    & \textbf{Success Rate} & $96.46\%$ & $96.13\%$ \\ 
    \cline{2-4}
    Complex 2D & \textbf{Average Samples Generated} & $29.75\pm27.22$ & $33.93\pm30.42$ \\ 
    \cline{2-4}
    & \textbf{Average Path Length} & $17.73\pm7.62$ & $17.48\pm7.01$\\ 
    \cline{2-4}
    & \textbf{Average Time} & $0.35\pm0.22$ s & $0.37\pm0.25$ s\\
    \hline
    \multirow{2}{*}{ Complex Rigid Body} & \textbf{Success Rate} & $95.63\%$ & $95.12\%$ \\ 
    \cline{2-4}
    & \textbf{Average Samples Generated}  & $30.16\pm29.63$ & $35.22\pm33.46$ \\ 
    \cline{2-4}
    & \textbf{Average Path Length} & $15.94\pm6.45$ & $15.58\pm6.33$ \\ 
    \cline{2-4}
    & \textbf{Average Time} & $0.38\pm0.25$ s & $0.42\pm0.27$ s \\ 
    \hline
    \multirow{2}{*}{Complex 3D} & \textbf{Success Rate} & $100\%$ & $100\%$  \\ 
    \cline{2-4}
    & \textbf{Average Samples Generated}  & $18.22\pm6.78$ & $17.97\pm7.27$ \\ 
    \cline{2-4}
    & \textbf{Average Path Length} & $20.56\pm5.01$ & $17.86\pm4.45$ \\ 
    \cline{2-4}
    & \textbf{Average Time} & $0.27\pm0.04$ s & $0.27\pm0.5$ s \\ 
    \hline
    \multirow{2}{*}{Snake} & \textbf{Success Rate} & $85.23\%$ & $86.53\%$  \\ 
    \cline{2-4}
    & \textbf{Average Samples Generated}  & $117.17\pm94.1$ & $106.23\pm87.29$ \\ 
    \cline{2-4}
    & \textbf{Average Path Length} & $15.21\pm5.96$ & $14.98\pm5.73$ \\ 
    \cline{2-4}
    & \textbf{Average Time} & $1.46\pm1.29$ s & $1.39\pm1.27$ s \\ 
    \hline
  \end{tabular}
  }
\end{table*}

In each workspace, we positioned 10 randomly placed obstacles of varying sizes to increase the complexity of the environments and to pose challenges to the SIL-RRT* algorithm.

For our training dataset, we employed the RRT* algorithm to collect feasible paths across various scenarios, amassing a total of 5000 data samples. Each sample represents a viable path in the designated environment.

To evaluate the effectiveness and generalization capabilities of SIL-RRT*, we created a comprehensive test dataset consisting of 100 workspaces, with each workspace featuring 10 unique scenarios. This dataset allows us to test SIL-RRT* on unseen scenarios, assessing its performance relative to other methodologies. The complex structure of the test dataset, including environments with an increased number of obstacles, serves to examine the scalability of SIL-RRT*.

In the test scenarios involving three distinct complex environments—2D, rigid body, and 3D—the number of obstacles was consistently set at 15. By subjecting SIL-RRT* to these enhanced obstacle conditions, we aim to evaluate its performance and gather insights into its ability to scale and effectively navigate through more challenging environments.

To further test SIL-RRT* in high-dimension scenarios, we designed a 2D space featuring a snake-shaped agent. This agent is composed of three links connected by two joints, creating a system with 5 degrees of freedom (DoF): the x and y positions of the first link, the rotation of the first link, and the angles of the two joints. To ensure the structural integrity of the agent, we restricted the joint angles to a range between -45 and 45 degrees, preventing the agent from folding upon itself. This setup allows us to explore the algorithm's capacity to handle high-dimensional movement and constraint scenarios effectively.

\subsection{Results}

In this study, we conducted a comprehensive performance benchmarking of SIL-RRT* against two well-known motion planning algorithms: RRT* and Motion Planning Networks with Neural Replanning (MPNetNR). Additionally, we compared the standard SIL-RRT* with an augmented version incorporating Weighted Self-Imitation Learning (WSIL) to assess the impact of WSIL on algorithmic performance. All algorithms were implemented in Python and evaluated using a specially curated dataset. Notably, the Deep Neural Network (DNN) used in the experiments was executed on a CPU.

The experimental suite, including training phases, was conducted on a high-performance computing system equipped with a 3.60 GHz Intel Core i9 processor and a NVIDIA GeForce RTX 3090 GPU. Both SIL-RRT* and MPNetNR were trained over 5000 iterations using a dataset where the number of point clouds representing obstacles was fixed at 1000. A distinction was made in the sampling method of the point clouds: SIL-RRT* samples were taken from the surfaces of obstacles, whereas MPNetNR samples were sourced from their interiors.

For testing in 2D and rigid body environments, the maximum number of samples per algorithm (RRT*, SIL-RRT*, MPNetNR) was set to 200. In contrast, for the 3D environments, which present increased complexity, the sample limit for RRT* was raised to 400, while SIL-RRT* and MPNetNR maintained a cap of 200 samples.

The goal region for all tested scenarios was uniformly set to 1. Importantly, the paths generated by these algorithms underwent no additional post-processing techniques such as path smoothing or lazy state contraction. This decision was made to ensure the evaluation of the algorithms' raw performance and facilitate a direct comparison in their original operational state. 

Fig.~\ref{fig:demos} presents search trees by SIL-RRT* and other planners for a selection of scenarios from the test dataset. As demonstrated, SIL-RRT* is capable of identifying near-optimal solutions across various environments with fewer nodes expanded.

Table~\ref{tab:comparison_simple} presents the performance of SIL-RRT* in scenarios across different environments, including 2D, rigid body, and 3D. The results are evaluated based on success rate, number of samples, path length, and computation time. To minimize the impact of probabilistic factors, we executed tree trials for each environment and calculated the mean and standard deviation of each property. The results demonstrate that SIL- RRT* outperforms RRT* and MPNetNR in terms of iden- tifying feasible solutions with reduced sample requirements and computation time across all environments. Furthermore, SIL-RRT* produces high-quality paths compared to the other algorithms.

We also conducted a comparison between SIL-RRT* using a pretrained model and one that underwent fine-tuning by WSIL. The results revealed that WSIL improved the quality of paths generated by SIL-RRT*, although it slightly in- creased the number of samples required.

MPNetNR showed a poor performance in the 2D and rigid body scenarios. This discrepancy is attributed to the size of our dataset, which is smaller than the dataset used in~\cite{qureshi2019motion}. Additionally, the density of obstacles in our scenarios may have posed additional challenges for MPNetNR.

These findings highlight the effectiveness of SIL-RRT* in generating high-quality paths while considering computational efficiency. The comparison with RRT* and MPNetNR underscores the advantages of our approach in scenarios with dense obstacles and limited dataset sizes.

To evaluate the scalability of SIL-RRT*, we conducted 4 additional trials on a more complex dataset. We utilized the same evaluation metrics as in the 10-obstacle dataset. The results of these trials are summarized in Table~\ref{tab:comparison_complex}
showing that our algorithm successfully handles more complex environments than those encountered during training. This finding highlights the adaptability and robustness of our method, as it performs well even in scenarios with increased environmental complexity.

\section{Conclusion and Limitation}

This study introduces SIL-RRT*, an innovative sampling-based algorithm developed to augment the efficiency of the Rapidly-exploring Random Tree (RRT) algorithm. Central to SIL-RRT* are two transformer architecture neural networks, serving as the sampler and estimator networks. These networks are responsible for generating samples and forecasting the length of solutions, respectively. To enhance SIL-RRT*'s performance, we implement a weighting mechanism that assigns values to each solution, predicated on the deviation between the actual solution length and its predicted counterpart. This weighting system is instrumental in assessing solution quality and mitigating the influence of suboptimal solutions within the framework of self-imitation learning.

The empirical investigations conducted across diverse environments and configurations unequivocally demonstrate SIL-RRT*'s superior performance relative to competing methodologies. The findings reveal that SIL-RRT* markedly enhances the convergence rate, diminishes the requisite number of samples, and facilitates the generation of solutions of superior quality. Collectively, these results position our proposed method as a notably promising solution to the motion planning quandary in robotics.

While SIL-RRT* has exhibited commendable performance using a constrained dataset and demonstrated the capacity to learn from generated paths, its dependency on substantial datasets for training remains a notable limitation. Procuring such extensive datasets becomes increasingly difficult in complex scenarios. As a prospective avenue for research, we propose exploring the integration of adaptive reinforcement learning (RL) techniques with the RRT* framework. This innovative approach would empower the algorithm to autonomously learn optimal paths, thereby obviating the need for extensive pre-collected datasets and potentially enhancing its applicability in diverse and intricate environments.

A notable limitation of SIL-RRT* lies in its presupposition of uniformity in the size and form of point-mass and rigid-body objects, a constraint that complicates applications involving robots of varying sizes or shapes. The algorithm's ability to generalize its learned knowledge to larger robots is, consequently, hindered. To surmount this challenge, our forthcoming research endeavors will involve the integration of robot size and shape considerations into our models. By incorporating these specific robot characteristics, we aim to significantly improve SIL-RRT*'s versatility in accommodating diverse robot configurations, thereby expanding its applicability in robotics.

\section{Acknowledgements}
The presented work has been supported by the Czech Science Foundation 
(GA\v{C}R) under the research project number 22-30043S.

\bibliographystyle{plain}
\bibliography{paper.bib}

\end{document}